\documentclass[journal]{IEEEtran}

\usepackage{multirow}

\ifCLASSINFOpdf
   \usepackage[pdftex]{graphicx}
\else
\fi
\usepackage{array}
\begin{document}

%
\title{Two Decades of Colorization and Decolorization for Images and Videos }
%
%
%

\author{Shiguang Liu
\thanks{Shiguang Liu is with College of Intelligence and Computing, Tianjin university, Tianjin 300350, P.R. China.
 (e-mail: lsg@tju.edu.cn).}
}

%
%

\markboth{}%
{Shell \MakeLowercase{\textit{et al.}}: Bare Demo of IEEEtran.cls for IEEE Journals}
%



\maketitle


\begin{abstract}
Colorization is a computer-aided process, which aims to give color to a gray image or video. It can be used to enhance black-and-white images, including black-and-white photos, old-fashioned films, and scientific imaging results. On the contrary, decolorization is to convert a color image or video into a grayscale one. A grayscale image or video refers to an image or video with only brightness information without color information. It is the basis of some downstream image processing applications such as pattern recognition, image segmentation, and image enhancement. Different from image decolorization, video decolorization should not only consider the image contrast preservation in each video frame, but also respect the temporal and spatial consistency between video frames. Researchers were devoted to develop decolorization methods by balancing spatial-temporal consistency and algorithm efficiency. With the prevalance of the digital cameras and mobile phones, image and video colorization and decolorization have been paid more and more attention by researchers. This paper gives an overview of the progress of image and video colorization and decolorization methods in the last two decades. 

\end{abstract}


%
\IEEEpeerreviewmaketitle

\section{Introduction}
%
%
%
%
\IEEEPARstart{I}mage manipulation is an important research topic in both computer graphics and image processing. Until now, many image manipulation topics have attracted attention from researchers, including image appearance transfer \cite{Liu2018Access, Liu2018AppearanceTansfer, Liu2012}, image synthesis \cite{Liu2011CAD, Cheng2017}, image analysis \cite{Hua2020, Liu2019Pose, Liu2019TIIS, Liu2015PRL}, image quality assessment \cite{Zhang2022TCSVT, Huang2021TMM, Huang2021TCSVT}, image2audio \cite{Liu2022Visual, Hao2021}, image segmentation \cite{Zhao2019, Zhao2020, Zhao2022IET, Zhao2022Electronics, Luo2014}, image inpainting \cite{Wei2016}, etc. This paper focuses on the techniques about colorization and decolorization.

There are a large number of gray-scale or black-and-white images and video materials in various film, television, picture archives, medical, and other fields. Coloring them can greatly enhance the detail features and help one better identify and use them. Traditional manual coloring method consumes a lot of manpower and material resources, and may not get satisfactory results. Given a source image or video, colorization methods aim to automatically colorize the target gray image or video reasonably and reliably, which thereby greatly improves the efficiency of this work.

Image or video decolorization, also known as grayscale transformation, converts a three-channel color image or video into a single-channel grayscale one. Decolorization is actually a process of \textit{dimension reduction}, so that the resulting grayscale image or video often only contains the most important information, which greatly saves storage space. A grayscale image or video can better display the texture and contour of objects. Decolorization can also be widely applied in the field of image compression, medical image visualization, and image or video art stylization. Black and white digital printing of images, with the advantages of low cost and fast printing, is common in daily life, one important process of which is decolorization, i.e., a color image sent to a monochrome printer must undergo a color-to-grayscale transformation.

Below we will summarize various image and video colorization and decolorization methods in the last two decades.

\section{Image Colorization}

 Colorization refers to adding colors to a grayscale image or video, which is a ill-posed task duet to that it is ambiguous to assign colors to a grayscale image or video without any prior knowledge. So, at the early stage, user intervention is usually involved in image colorization. Later, automatic image colorization methods and deep-learning based colorization methods emerged. 
 
 \subsection{Semi-Automatic Colorization}

Semi-automatic colorization methods require some amounts of user interactions. Among them, color transfer methods (\cite{Welsh2002, Reinhard2001, Abadpour2007} and image analogy methods (\cite{Hertzmann2001} in Chapter 2 are widely used. In this case, a source image is provided as an example for coloring a given grayscale image, i.e., the target image. When the source image and the grayscale image share similar contents, impressive colorization results can be achieved. Nevertheless, these methods are labor intensive, since the source image and the target image should be manually matched.

A luminance keying based method for transferring color to a grayscale image is described in Gonzalez and Woods \cite{Gonzalez1987}. Color and grayscale values are matched with a pre-defined look-up table. When assigning different colors for a same gray level, a few luminance keys should be simultaneously manipulated by the user for different regions, making it a tiresome process. As an extension of color transferring method between color images \cite{Reinhard2001}, Welsh et al. \cite{Welsh2002} proposed to transferring from a source color image to a target grayscale image. It matches color information between the two images with swatches. 

\begin{figure}[t] 
\centering
  \includegraphics[width=3.5in]{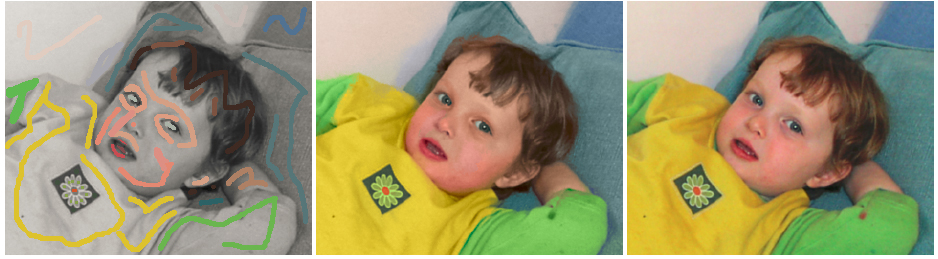}
\caption{Illustration of the colorization method using optimization \cite{Levin2004}. From left to right: an input grayscale image marked with color scribbles by the user, the colorization result by \cite{Levin2004} (middle), and the ground truth.}
\label{ch04.fig1} 
\end{figure}

Levin et al. \cite{Levin2004} presented an efficient colorization method which allows users to interact with a few scribbles. With the observation that neighboring pixels in space-time share similar intensities should have similar colors, they formulate colorization as an optimization problem for a quadratic cost function. As shown in Fig. \ref{ch04.fig1}, with a few color scribbles by the user, the indicated colors can be automatically propagated in the grayscale image. Nie et al. \cite{Nie2005} developed a colorization method by a local correlation based optimization algorithm. This method depends on the color correlativity between pixels in different regions, limiting its practical application. Nie et al. \cite{Nie2007} presented an efficient grayscale image colorization method. This method achieved comparable colorization quality with \cite{Levin2004} with much less computation time by quadtree decomposition based non-uniform sampling. Furthermore, this method greatly reduces the problem of color diffusion among different regions via designing a weighting function to represent intensity similarity in the cost function. This is an interactive colorization method, where the user need provide color hints by scribbling or seed pixels.

Irony et al. \cite{Irony2005} presented a novel colorization method by transferring color from an exampler image. This method uses a strategy of higher-level context of each pixel instead of independent pixel-level decisions in order to achieve better spatial consistency in the colorization result. Specifically, with a supervised classification scheme, they estimate the best example segment for each pixel to learn color from. Then, by combining a neighborhood matching metric and a spatial filter for high spatial coherence, each pixel is assigned a color from the corresponding region in the example image. It is reported that this approach requires considerably less scribbles than previous interactive colorization methods (e.g., \cite{Levin2004}).

\begin{figure}[t] 
\centering
  \includegraphics[width=3.5in]{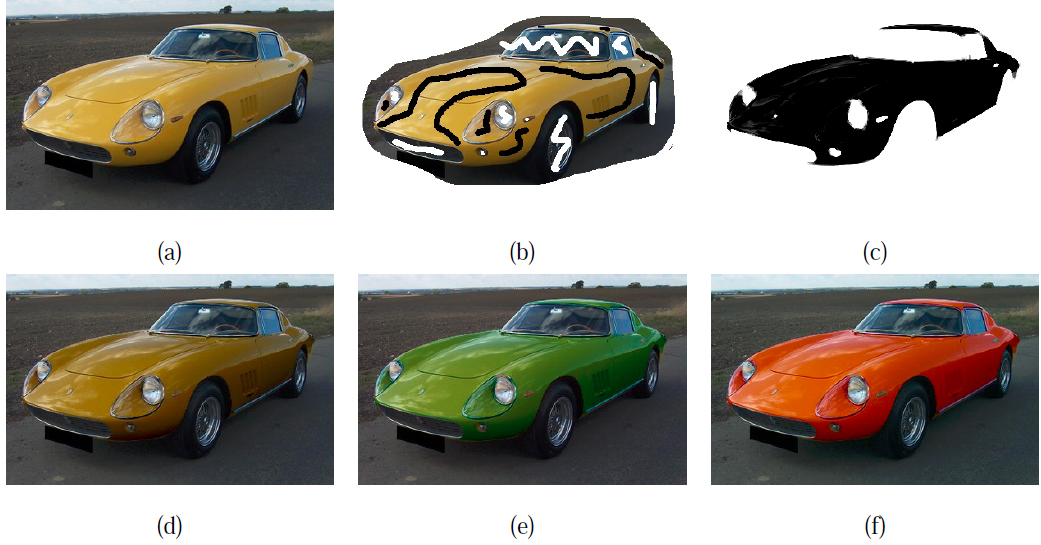}
\caption{The application of the chrominance blending based colorization method for image recolorization \cite{Yatziv2006}. }
\label{ch04.fig3} 
\end{figure}

Yatziv and Sapiro \cite{Yatziv2006} proposed an image colorization method via chrominance blending. This scheme is based on the concept of color blending derived from a weighted distance function that is computed from the luminance channel. This method is fast and allows the user to interactively colorize a grayscale image by providing a reduced set of chrominance scribbles. This method can also be extended for recolorization and brightness change (Fig. \ref{ch04.fig3}). As shown in Fig. \ref{ch04.fig3}, given a target color image (a), the goal is to recolorize the yellow car into a darker one. Firstly, the blending medium is defined by simply marking areas to be changed and unchanged with scribbles (b). Then, the marks are propagated to be a gray-scale matte (c). The brightness of the target image is changed by subtracting the grey-level matte from the intensity channel (d). (e) and (f) show more recolorization results by adding the grey-level matte to the $C_b$ and $C_r$ channels, respectively. 

\begin{figure}[t] 
\centering
  \includegraphics[width=3.5in]{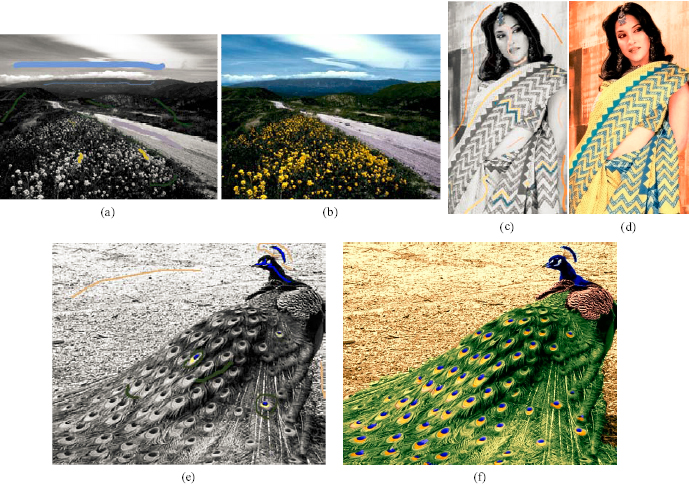}
\caption{Results of colorizing texture with multiple colors \cite{Luan2007Natural}. Note that the strokes are shown in (a), (c), and (e), and the corresponding colorization results are presented in (b), (d), and (f). Colorizing texture with multiple colors is the unique property of this method.}
\label{ch04.fig4} 
\end{figure}

Image color can be viewed as a highly correlated vector space. Abadpour and Kasaei \cite{Abadpour2007} realized grayscale colorization by applying the PCA (Principal Component Analysis) based transformation. They propose a category of colorizing methods that generate the color vector corresponding to the grayscale as a function. This method is significantly faster than previous approaches while producing visually acceptable colorization results. It can also be extended for recolorization. Nevertheless, this method is restricted by complicate segmentation that is tiresome by using the \textit{magic select} tool in \textit{Adobe Photoshop}. Luan et al. \cite{Luan2007Natural} proposed an interactive system for users to easily colorize natural images. The colorization procedure consists of two stages: color labeling and color mapping. In the first stage, pixels that should have similar colors are grouped into coherent regions in the first stage, while in the second stage color mapping is applied to generate vivid colorization effect through assigning colors to a few pixels in the region. It is very tedious to colorize texture by previous methods since each tiny region inside the texture need a new color. In contrast, this method handles texture by grouping both neighboring pixels sharing similar intensity and remote pixels with similar texture (see Fig. \ref{ch04.fig4}). This method is effective for natural image colorization. However, the user should usually provide multiple stokes on similar patterns with different orientation and scales in order to produce fine colorization results.

Liu et al. \cite{Liu2008} proposed an example-based colorization method that is aware of illumination differences between the target grayscale image and the source color image. Firstly, an illumination-independent intrinsic reflectance map of the target scene is recovered from multiple color references collected by web search. Then, the grayscale versions of the reference images are employed for decomposing the target grayscale image into its intrinsic reflectance and illumination components. The color is transferred from the color reflectance map to the grayscale reflectance image. By relighting with the illumination component of the target image, the final colorization result can be produced. This method needs to search suitable source images for reference by web search.

Liu et al. \cite{Liu2011} presented a gray-scale image colorization method by control of single-parameter. The polynomial fitting model of the histograms of the source image and the gray-scale image are computed by linear regression, respectively. With the user-assigned order of the polynomials, the source image and the grayscale images can be automatically matched. By transferring between the corresponding regions of the source image and the gray-scale image, colorization can be finally achieved. Quang et al. \cite{Quang2010} proposed an image and video colorization method based on the kernel Hilbert space (RKHS). This method can produce impressive colorization results. Nevertheless, it requires initialization for different regions by manual, that is time-consuming if there are many different contents in the grayscale image.

\subsection{Automatic Colorization}
 The above colorization methods require the user to perform colorization by manual, either providing a source image or using scribbles and color seeds for interaction. Since there is usually no suitable correspondence between color and local texture, automatic colorization is necessary. 
 
 Li and Hao \cite{Li2008} proposed an automatic colorization approach by locally linear embedding. Given a source color image and a target grayscale image, this method clips both of them into overlapping patches, which are supposed to be distributed on a manifold \cite{Fan2007, Beymer1996}. For each patch, its neighborhood in the training patches is estimated and its chromatic information is predicted by the manifold learning \cite{Roweis2000}. With multimodality, Charpiat et al. \cite{Charpiat2008} predict the probability distribution of all possible colors for each pixel of the image to be colored,, rather than selecting the probable color locally. Then, the technique of graph cut is employed to maximize the probability of the whole colored image globally. Morimoto et al. \cite{Morimoto2009} proposed an automatic colorization method using multiple images collected from the web. Firstly, this method chooses images with similar scene structure with the target grayscale image $I_m$ as the source images. The \textit{gist scene descriptor}, a feature vector expressing the global scene in a lower dimension is used to aggregate oriented edge responses at multi-scales into coarse spatial bins. Then the distance between the gist of $I_m$ and that of the images from the web are computed. The most similar images are chosen as source images, which are used for colorization. Here, the transferring method of \cite{Welsh2002} was used for colorization. However, this method restricts from the searching results from the images collected from the web, which may produce unnatural results due to the source images that are structurally similar but semantically different. 

\begin{figure}[t] 
\centering
  \includegraphics[width=3.6in]{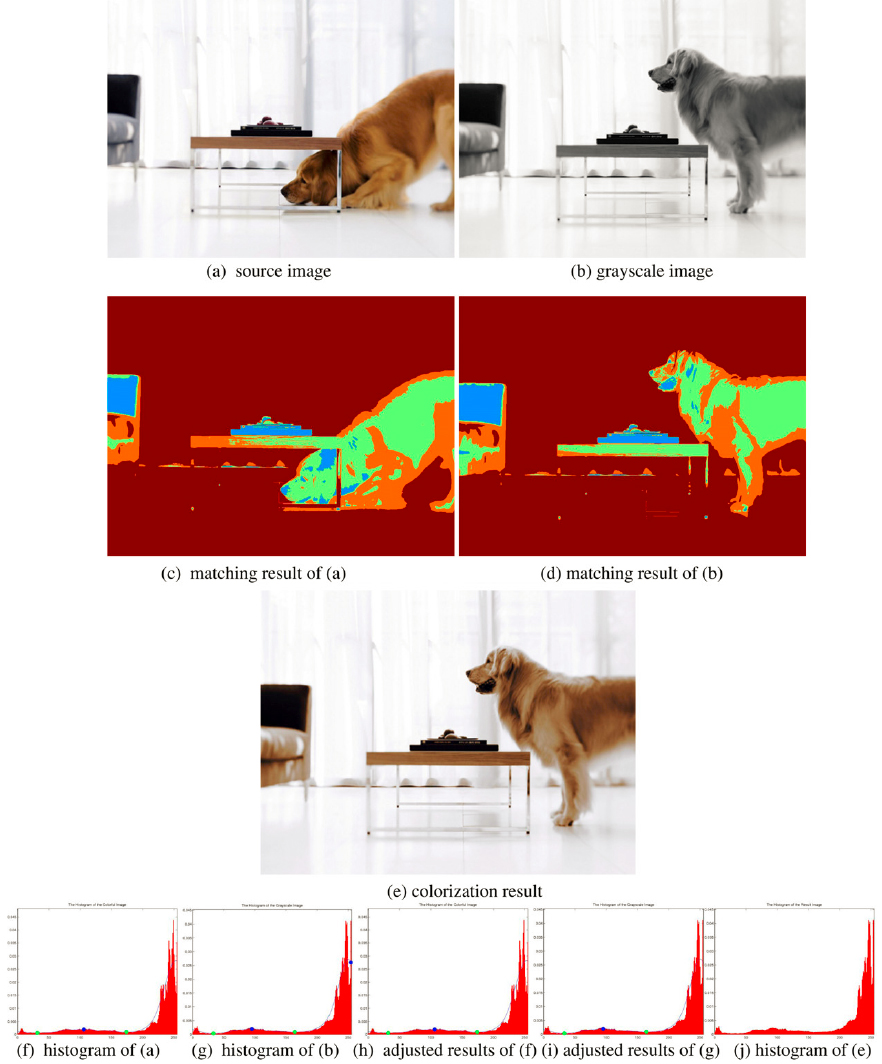}
\caption{A colorization result of an animal image using the automatic grayscale image colorization method via histogram regression \cite{Liu2012Colorization}.}
\label{ch04.fig7} 
\end{figure}

To this end, Liu and Zhang \cite{Liu2012Colorization} proposed an automatic grayscale image colorization method via histogram regression. Given a source image and a target grayscale image, the locally weighted regression is performed on both images to obtain the feature distributions of them. Then, these features are automatically matched by aligning zero-points of the histogram. Thus, the grayscale image is colorized in a weighted manner. Figure \ref{ch04.fig7} shows a colorization result by this method. Although this method can achieve confident colorization results, it may fail for images with strong texture patterns or varied lighting effects (e.g., shadows and highlight). 

Liu and Zhang \cite{Liu2013} further proposed a colorization method based on texture map. Assuming that a source color image with the similar content with the target grayscale image can be provided by the user, this method is aware of both the luminance and texture information of images so that more convincing colorization results can be produced. Specifically, given a source color image and a target grayscale image, their respective spatial maps are computed. Note that the spatial map is a function of the original image, indicating the luminance spatial distribution for each pixel. Then, by performing locally weighted linear regression on the histogram of the quantized spatial map, a series of spatial segments are computed. Within each segment, the luminance of target grayscale image is automatically mapped to color values. Finally, colorization results can be yielded through local luminance-color correspondence and global luminance-color correspondence between the source color image
and the target grayscale image.

Beyond natural images, Visvanathan et al. \cite{Visvanathan2007} automatically colorized pseudocolor images by gradient-based value mapping. This method targets for visualizing pixel values and their local differences for scientific analysis.

\begin{figure}[t] 
\centering
  \includegraphics[width=3.5in]{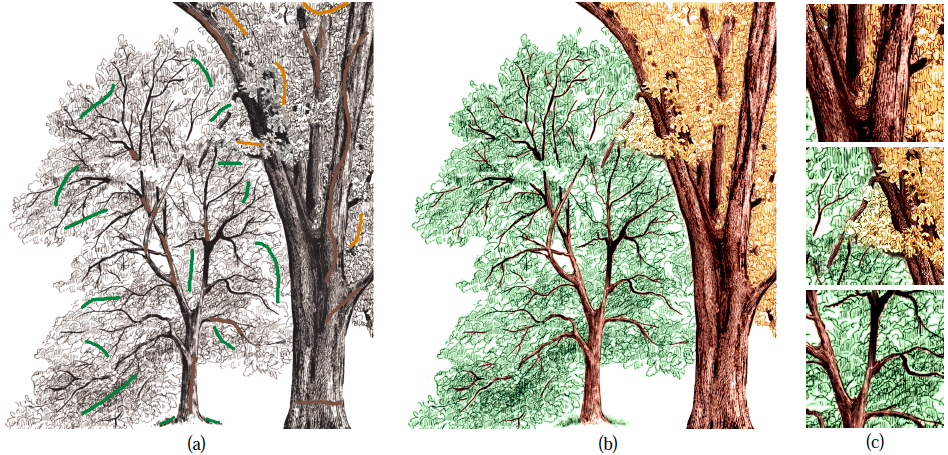}
\caption{An example of stroke-preserving manga colorization \cite{Qu2006}. (a) the target manga drawing with user scribbles, (b) the colorization result, and (c) the enlarged views. Note that a color-bleeding algorithm is utilized here, so that even if the user provides careless scribbles, the leave region can still be accurately separated from tree branches.}
\label{ch04.fig8} 
\end{figure}

\subsection{Cartoon Colorization} 

Some researchers also extended the colorization technique to cartoon images. S\'{y}kora et al. \cite{Sykora2003} proposed a semi-automatic, fast and accurate segmentation method for black and white cartoons. It allows the user to efficiently apply ink on the aged black and white cartoons. The inking process is composed of four stages, namely segmentation, marker prediction, color luminance modulation, and final composition of foreground and background layers.

Qu et al. \cite{Qu2006} proposed a method for colorizing black-and-white manga (comic books in Japanese) containing a large number of strokes, hatching, halftoning, and screening. Given scribbles by the user on the target grayscale manga drawing, Gabor wavelet filters are employed to measure the pattern-continuity and thereby a local, statistical based pattern feature can be estimated. Then, with the level set technique, the boundary is propagated to monitor the pattern continuity. In this way, areas with open boundaries or multiple disjointed regions with similar patterns can be well segmented. Once the segmented regions are obtained, conventional colorization methods can be used to color replacing, color preservation as well as pattern shading. Figure \ref{ch04.fig8} shows an example of stroke-preserving manga colorization by this method. 

\subsection{Deep Colorization}

Cheng et al. \cite{Cheng2015} proposed a deep neural network model to achieve fully automatic image colorization by leveraging a large set of source images from different categories (e.g., animal, outdoor, indoor) with various objects (e.g., tree, person, panda, and car). This method consists of two stages, (1) training a neural network, and (2) colorizing a target grayscale image with the learned neural network.

\begin{figure}[t] 
\centering
  \includegraphics[width=3.5in]{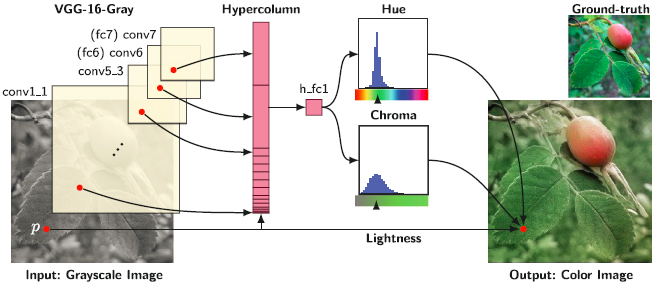}
\caption{The framework of the automatic colorization method via learning representations \cite{Larsson2016}.}
\label{ch04.fig9} 
\end{figure}

Larsson et al. \cite{Larsson2016} trained a model to predict per-pixel color histograms for colorization. This method trains a neural architecture in an end-to-end manner by considering semantically meaningful features of varying complexity. Then, a color histogram prediction framework is developed to treat uncertainty and ambiguities inherent in colorization so as to avoid jarring artifacts. As shown in Fig. \ref{ch04.fig9}, given a grayscale image, with a deep convolutional architecture (VGG), spatially localized multilayer slices are chosen as per-pixel descriptors. The system then estimates hue and chroma distributions for each pixel $p$ with its hypercolumn
descriptor. Finally, at test time, the estimated distributions are used for color assignment. 

Zhang et al. \cite{Zhang2016} treat image colorization as a classification problem considering the underlying uncertainty of this task. They leverage class-rebalancing during training to increase the diversity of colors. At test time, this method is performed as a feed-forward pass in a CNN with a million color images. This method demonstrates that with a deep CNN and a carefully-tuned loss function, the colorization task can generate results close to real color photos.

\begin{figure}[t] 
\centering
  \includegraphics[width=3.5in]{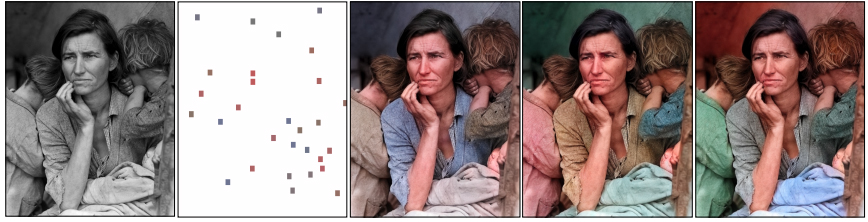}
\caption{A result of the real-time user-guided deep colorization method with learned deep priors \cite{Zhang2017}. Given a target grayscale image (the left image) and sparse user edits (the second left image), multiple plausible colorization results (the right three images). Photograph of Migrant Mother by Dorothea Lange, 1936 (Public Domain).}
\label{ch04.fig10} 
\end{figure}

Iizuka et al. \cite{Iizuka2016} proposed an automatic, CNN-based grayscale image colorization method by combining both global priors and local image features. The proposed network architecture is able to jointly extract global and local features from an image and fuse them for colorization. Specifically, their model is composed of four parts, namely a low-level features network, a mid-level features network, a global features network, and a colorization network. Various evaluation experiments were performed to verify this method with user study and many historical hundred-year-old black-and-white photographs. Figure \ref{ch04.fig10} shows an example of this method.

Zhang et al. \cite{Zhang2017} propose a CNN framework for user-assisted image colorization. Given a target grayscale image, and sparse, local user edits, this method can automatically produce convincing colorization results. By training on a large amount of image data, this method learns to propagate user edits by merging both low-level cues and high-level semantic information. This method has help non-professionals to design a colorful work, since it has great ability to achieve fine colorization results even with random user inputs. 

Deshpande et al. \cite{Deshpande2017} learned a low-dimensional smooth embedding of color fields with a variational autoencoder (VAE) for grayscale image colorization. A multi-modal conditional model between the gray-level features and the low-dimensional embedding is learned to produce \textit{diverse} colorization results. The loss functions are specially designed for the VAE decoder to avoid blurry colorization results and respect uneven distribution of pixel colors. This method has potential to handle other ambiguous problems, since the low-dimensional embeddings has ability to predict diversity with multi-modal conditional models. However, high spatial detail is not taken into account in this method.

\section{Image Decolorization}

Image decolorization is often used as a preprocessing for downstream image processing tasks such as segmentation, recognition, and analysis. Recently, decolorization has attracted more and more attention of researchers. In the early stage, the three channels $R$, $G$, and $B$ are represented by a single channel or only the brightness channel information is used to represent the grayscale image. However, these simple color removal methods suffer from contrast loss in the gray image. To this end, researchers have proposed local and global decolorization methods in order to preserve the contrast of color images in the resulting grayscale images.

\subsection{Early Decolorization Methods}

The early image decolorization method is simple, which directly processes the $(R, G, B) $ channels of a color image in the $RGB$ color space. These methods include the component method, the maximum method, the average method, and the weighted average method.

\textit{The component method} uses one of the $(R,G,B)$ in the color image as the corresponding pixel value in the grayscale image, written as
\begin{equation}
\begin{array}{l}
{G_1}(i,j) = R(i,j),\\
{G_2}(i,j) = G(i,j),\\
{G_3}(i,j) = B(i,j),
\end{array}
\end{equation}
where $(i,j)$ is the pixel coordinate in an image. Note that any one of ${G_1},{G_2},{G_3}$ can be selected as needed.

\textit{The maximum method} takes the maximum value of $(R, G, B)$ in the color image as the gray value of the grayscale image.
\begin{equation}
GRAY(i,j) = \max \{ R(i,j),G(i,j),B(i,j)\}.
\end{equation}

\textit{The average method} is to average the three component values of $(R, G, B) $ in the color image to obtain a gray value.
\begin{equation}
GRAY(i,j) = (R(i,j),G(i,j),B(i,j))/3.
\end{equation}

\textit{The weighted average method} uses the weighted average of three components with different weights as the grayscale image.
\begin{equation}
GRAY(i,j) = 0.299 R(i,j) + 0.578 G(i,j) + 0.114 B(i,j).
\end{equation}

In addition to using the color component of the $RGB$ space, it is also common to employ the brightness channel of other color spaces to represent the gray value of a grayscale image. For example, Hunter \cite{Hunter1958} uses the $L$ channel of the $L\alpha\beta$ space to represent a grayscale image, while Wyszecki and Stiles \cite{Wyszecki1968} adopt the $Y$ component in the $YUV$ color space to represent the grayscale image. In the $YUV$ color space, the $Y$ component is the brightness of pixels, reflecting the brightness level of an image. According to the relationship between the $RGB$ color space and the $YUV$ color space, the mapping between the brightness $y$ and three color components can be established as
\begin{equation}
y = 0.3r + 0.59g + 0.11b.
\end{equation}
The luminance value $y$ is used to represent the gray value of the image. Based on this observation, Nayatani \cite{Nayatani2015} proposed a color mapping model with independent input, i.e., input three components independently and set the weights of the corresponding components as needed.

These early methods are easy to implement, however, they would cause the loss of image contrast, saturation, exposure, etc. To this end, researchers explored decolorization methods with higher accuracy and efficiency, including local decolorization methods, global decolorization methods, and deep learning based decolorization methods.

\subsection{Local Decolorization Methods}

Local decolorization methods usually use different strategies in solving the mapping model from a color image to a grayscale one. The strategy deals with different pixels or color blocks, and increases the local contrast by strengthening the local features.

\begin{figure}[t] 
\centering
  \includegraphics[width=3.5in]{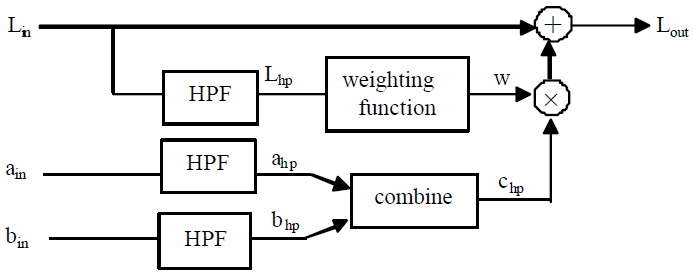}
\caption{The flow chart of the spatial color-to-grayscale transform method \cite{Bala2004}. Here the $Lab$ color space is taken as an example. Note that "HPF" represents high-pass filter.}
\label{ch05.fig1} 
\end{figure}

Bala and Eschbach \cite{Bala2004} proposed a decolorization method that locally enhances the edge and contours between adjacent colors through adding high-frequency chrominance information into the luminance channel. Specifically, a spatial high-pass filter weighting the output with a luminance-dependent term is applied to the chrominance channels. Then the result is added to the luminance channel. Figure \ref{ch05.fig1} shows a flow chart of this method. 

\begin{figure}[t] 
\centering
  \includegraphics[width=3.5in]{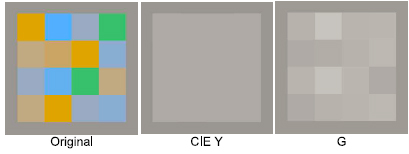}
\caption{A decolorization result showing that isoluminant colors can be mapped to unique, properly ordered gray values \cite{Smith2008}.}
\label{ch05.fig2} 
\end{figure}

Neumann et al. \cite{Neumann2007} view the color and luminance contrasts of an image as a gradient field and solve the inconsistency of the field. They chose locally consistent color gradients and performed 2D integration to produce the grayscale image. Since its complexity is linear in the number of pixels, this method is simple yet very efficient, which is suitable for handling high-resolution images. Smith et al. \cite{Smith2008} proposed a perceptually accurate decolorization method for both images and videos. This approach consists of two steps: (1) globally assigning gray values and determine color ordering, and (2) locally improving the grayscale to preserving the contrast in the input color image. The Helmholtz-Kohlrausch color appearance effect is introduced to estimate distinctions between isoluminant colors. They also designed a multiscale local contrast enhancement strategy to produce a faithful grayscale result. Note that this method makes a good balance between a fully automatic method (first step) and user assist (second step), making it suitable for dealing with various images (e.g., natural images, photographs, artistic works, and business graphics). Figure \ref{ch05.fig2} shows that, for a challenging image consists of equiluminant colors, this method is able to predict the H-K effect that makes a more colorful blue appear lighter than the duller yellow. A limitation of this approach comes from the locality of the second step, which may fail to preserve chromatic contrast between non-adjacent regions and lead to temporal inconsistencies.

Lu et al. \cite{Lu2012} proposed a decolorization method aiming to preserving the original color contrast as far as possible. A bimodal contrast preserving function is designed to constrain local pixel differences and a parametric optimization approach is employed to preserve the original contrast. Owing to weak color order constraint, they relax the color order constraint and seek to better maintain color contrast and enhance the visual distinctiveness for edges. Nevertheless, this method cannot greatly preserve the global contrast in the image. Moreover, since the gray image is produce by solving the energy equation in an iterative manner, the efficiency of this algorithm is relatively low. Zhang and Liu \cite{Zhang2017Decolorization} presented an efficient image decolorization method via perceptual group difference (PGD) enhancement. They view the perceptual group instead of individual image pixels as the human perception elements. Based on this observation, they perform decolorization for different groups in order to maximumly maintain the contrast between different visual groups. A global color to gray mapping is employed to estimate the grayscale of the whole image. Experimental results showed that, with PGD enhancement, this approach is capable of achieving better visual contrast effects.

The local decolorization methods may distort appearance for regions with constant colors and therefore lead to undesired haloing artifacts.

\subsection{Global Decolorization Methods}
Global decolorization methods perform decolorization on the whole image in a global manner, including linear declorization and nonlinear decolorization techniques.

\textbf{Linear declorization methods.} Gooch et al. \cite{Gooch2005} proposed Color2Gray, a saliency-preserving decolorization method. This method is performed in the CIE $L^*a^*b^*$ color space instead of the traditional $RGB$ color space. Considering that the human visual system is sensitive to change, they preserve relationships between neighboring pixels rather than representing absolute pixel values. The chrominance and luminance changes in a source image are transferred to changes in the target grayscale image so as to produce images maintaining the salience of the source color images. Grundland and Dodgson \cite{Grundland2007} proposed an efficient, linear decolorization approach by adding a fixed amount of chrominance to lightness. To achieve a perceptually plausible decolorization result, Kuk et al. \cite{Kuk2010} proposed a color to grayscale conversion method by taking into account both local and global contrast. They encode both local and global contrast into an energy function via a target gradient field, which is constructed from two types of edges: (1) edges connecting each pixel to neighboring pixels, and (2) edges connecting each pixel to predetermined landmark pixels. Finally, they formulate the decolorization problem as reconstructing a grayscale image from the gradient field, that is solved by a fast 2D Poisson solver. 

\begin{figure}[t] 
\centering
  \includegraphics[width=3.5in]{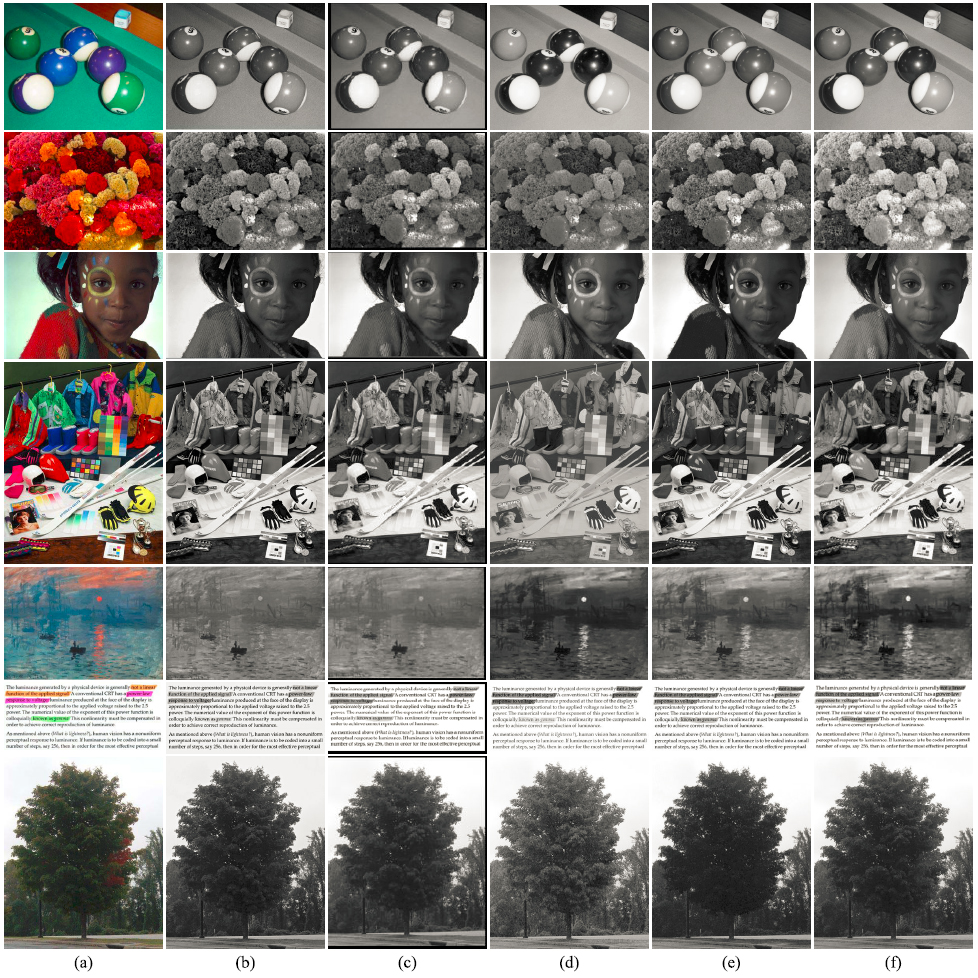}
\caption{A comparison on the dataset \cite{Cadik2008} among different methods \cite{Liu2015GcsDecolor}. (a) through (f) are the input color image, the decolorization result of Smith et al. \cite{Smith2008}, the decolorization result of Kim et al. \cite{Kim2009}, Kim et al. \cite{Kim2009}, the decolorization result of Lu et al. \cite{Lu2012}, the decolorization result of Lu et al. \cite{Lu2012Asia}, and the decolorization result of Liu et al. \cite{Liu2015GcsDecolor}.}
\label{ch05.fig3} 
\end{figure}

\textbf{Nonlinear declorization methods.} 
Kim et al. \cite{Kim2009} presented a fast and robust decolorization algorithm via a global mapping that is a \textit{nonlinear} function of the lightness, chroma, and hue of colors. Given a color image, the parameters of the function are optimized to make resulting grayscale image respect the feature discriminability, lightness, and color ordering in the input color image. Ancuti et al. \cite{Ancuti2010} introduced a fusion-based decolorization technique. The input of their method include three independent $RGB$ channels and an additional image that conserves the color contrast. The weights are based on three different forms of local contrast: a saliency map to preserve the saliency of the original color image, a second weight map taking advantages of well-exposed regions, and a chromatic weight map enhancing the color contrast. By enforcing a more consistent gray-shades ordering, this strategy can better preserve the global appearance of the image. Ancuti et al. \cite{Ancuti2011} further presented a color to gray conversion method aiming to enhance the contrast of the images while preserving the appearance and quality in the original color image. They intensify the monochromatic luminance with a mixture of saturation and hue channels in order to respect the original saliency while enhancing the chromatic contrast. In this way, a novel spatial distribution can be produced which is capable of better discriminating the illuminated regions and color features. Liu et al. \cite{Liu2015GcsDecolor} developed a decolorization model based on gradient correlation similarity (Gcs) so as to reliably maintain the appearance of the source color image. The gradient correlation is employed as a criterion to design a \textit{nonlinear} global mapping in the $RGB$ color space. Figure \ref{ch05.fig3} shows a comparison result between this method and other image decolorization methods including Smith et al. \cite{Smith2008}, Kim et al. \cite{Kim2009}, Lu et al. \cite{Lu2012}, and Lu et al. \cite{Lu2012Asia}. It can be seen from the results that this method is able to better preserve features in the source color image which are more discriminable in the grayscale image; and it also has good ability to maintain a desired color ordering in color-to-gray conversion. Liu et al. \cite{Liu2016RGB2Gray} further proposed a color to grayscale method by introducing the gradient magnitude \cite{Xue2014}.

\begin{figure}[t] 
\centering
  \includegraphics[width=3.5in]{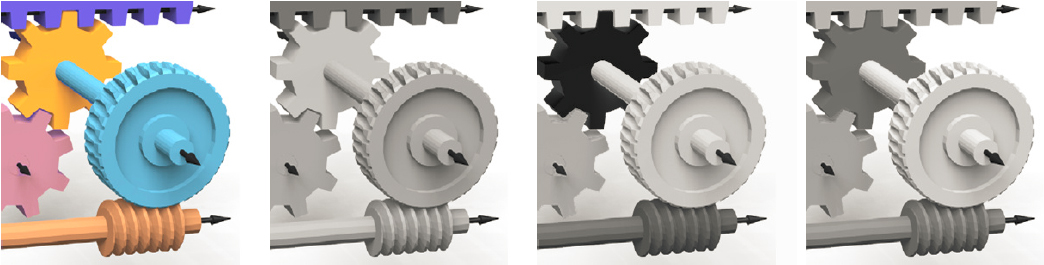}
\caption{A comparison among different decolorization result with running time \cite{{Lu2012Asia}}. From left to right: the input color image, the result of the built-in Matlab function \textit{rgb2gray} (8 $ms$), the result of \cite{Lu2012} (1,102 $ms$), and the result of \cite{{Lu2012Asia}} (30 $ms$). Note that all the methods were implemented in Matlab.}
\label{ch05.fig4} 
\end{figure}

Song et al. \cite{Song2017Decolorization} regard decolorization as a labeling problem to maintain the visual cues of a color image in the resulting grayscale image. They define three types of visual cues, namely color spatial consistency, image structure information, and color channel perception priority, that can be extracted from a color image. Then, they cast color to gray as a visual cue preservation process based on a probabilistic graphical model, which are solved via integral minimization. 

Most of the above image decolorization methods attempt to preserve as much as possible visual appearance and color contrast, however, little attention was devoted to the speed issue of decolorization. The efficiency of most method is lower than the standard procedure (e.g., Matlab built-in \textit{rgb2gray} function). To this end, Lu et al. \cite{Lu2012Asia} proposed a \textit{real-time} contrast preserving decolorization method. They achieved this goal by three main ingredients: a simplified bimodal objective function with linear parametric grayscale model, a fast non-iterative discrete optimization, and a sampling based $P$-shrinking optimization strategy. The running time of this method is a constant $O(1)$, independent of image resolutions. As shown in Fig. \ref{ch05.fig4}, this method takes only 30ms (the rightmost result) to decolorize an one megapixel color image, that is comparable with the built-in Matlab \textit{rgb2gray} function (the left second result), but achieving a better color to gray conversion result which is visually similar to a compelling contrast preserving decolorization method \cite{Lu2012} (the right second result). 

\begin{figure}[t] 
\centering
  \includegraphics[width=3.5in]{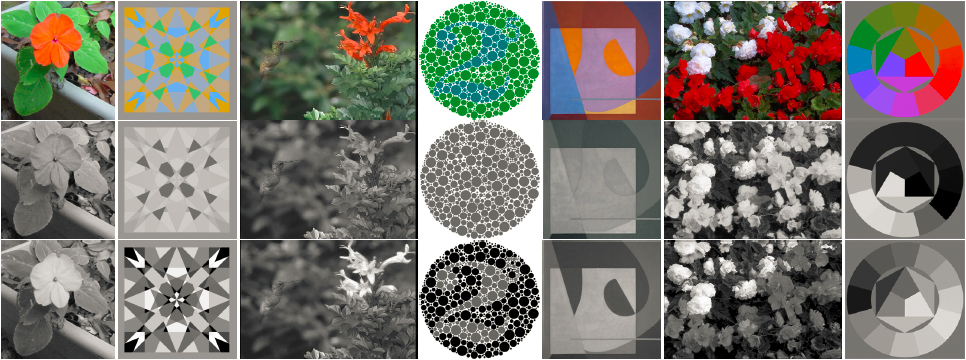}
\caption{A comparison of the robust decolorization method by Song et al. with conventional methods \cite{Song2013}. The top row is the original color images. The middle row shows the failure results of current decolorization methods (from left to right: Gooch et al. \cite{Gooch2005}, Gundland and Dodgson \cite{Grundland2007}, \cite{Smith2008}, Kim et al. \cite{Kim2009}, Ancuti et al. \cite{Ancuti2011}, Lu et al. \cite{Lu2012}, and Lu et al. \cite{Lu2012Asia}). The bottom row are results by Song et al. \cite{Song2013} which are produced by modifying \textit{rgb2gray()} with adjusted weights for $R$, $G$, and $B$ channels.}
\label{ch05.fig5} 
\end{figure}

Lu et al. \cite{Lu2014} further presented an optimization framework for image decolorization to preserve color contrast in the original color image as much as possible. A bimodal objective function is used to reduce the restrictive order constraint for color mapping. Then, they design a solver to realizing automatic selection of suitable grayscales via global contrast constraints. They also propose a quantitative perceptual-based metric, E-score, to measure contrast loss and content preservation in the resulting grayscale images.  The E-score is to jointly consider two measures CCPR (Color Contrast Preserving Ratio) and CCFR (Color Content Fidelity Ratio), written as 
\begin{equation}
    E_{score}= \frac {2 \cdot {CCPR} \cdot {CCFR}} {{CCPR} + {CCFR}}
\end{equation}
It is reported that this is among the first attempts in the color to gray field to quantitatively evaluate decolorization results.

Considering that the above decolorization methods suffer from the robustness problem, i.e., may fail to accurately convert iso-luminant regions in the original color image, while the $rgb2gray()$ function in Matlab works well in practice applications. Song et al. \cite{Song2013} proposed a robust decolorization method by modifying the \textit{rgb2gray()} function. Figure \ref{ch05.fig5} shows that this method is able to realize color to gray conversion for iso-luminance regions in an image, while previous methods, including Gooch et al. \cite{Gooch2005}, Gundland and Dodgson \cite{Grundland2007}, \cite{Smith2008}, Kim et al. \cite{Kim2009}, Ancuti et al. \cite{Ancuti2011}, Lu et al. \cite{Lu2012}, and Lu et al. \cite{Lu2012Asia} fail in this task. In this method, they avoid indiscrimination in iso-luminant regions by adaptively selecting channel weights with respect to specific images rather than using fixed channel weights for all cases. Therefore, this method is able to maintain multi-scale contrast in both spatial and range domain.

Sowmya et al. \cite{Sowmya2016} presented a color to gray conversion algorithm with a weight matrix corresponding to the chrominance components. The weight matrix is obtained by reconstructing the chrominance data matrix through singular value decomposition (SVD). Ji et al. \cite{Ji2016} presented a global image decolorization approach with a variant of difference-of-Gaussian band-pass filter, called luminance filters. Typically, the filter has high responses on regions of which colors differ from their surroundings for a certain band. Then, the grayscale value can be produced after luminance passing a series of band-pass filters. Due to that this approach is linear in the number of pixels, it is efficient and easy to implement,

\subsection{Deep Learning Based Decolorization Methods}

By training partial differential equations (PDEs) on 50 input/output image pairs, Lin et al. \cite{Lin2008} constructed a mapping model for the task of color to gray conversion. It is reported that their learned PDEs can yield similar decolorization results with those of Gooch et al. \cite{Gooch2005}.

\begin{figure}[t] 
\centering
  \includegraphics[width=3.5in]{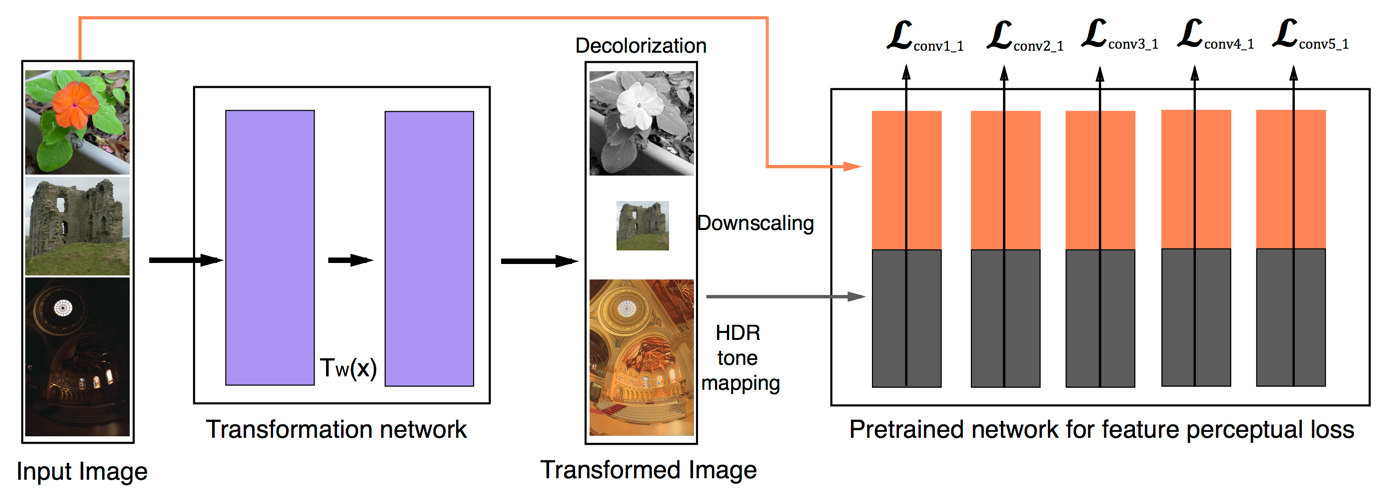}
\caption{An overview of the deep feature consistent deep image transformation (DFC-DIT) framework \cite{Hou2017}.}
\label{ch05.fig6} 
\end{figure}

Hou et al. \cite{Hou2017} proposed the Deep Feature Consistent Deep Image Transformation (DFC-DIT) framework for one-to-many mapping image processing tasks (e.g., downscaling, decolorization, and tone mapping). The DFC-DIT achieves transformation between images with a CNN as a non-linear mapper respecting the deep feature consistency principle that is enforced with another pretrained and fixed deep CNN. As shown in Fig. \ref{ch05.fig6}, this system is comprised of two networks, a transformation network and a loss network. The former is used to convert an input to an output, and the later servers as computing the feature perceptual loss for the training of the transformation network.

\begin{figure}[t] 
\centering
  \includegraphics[width=3.5in]{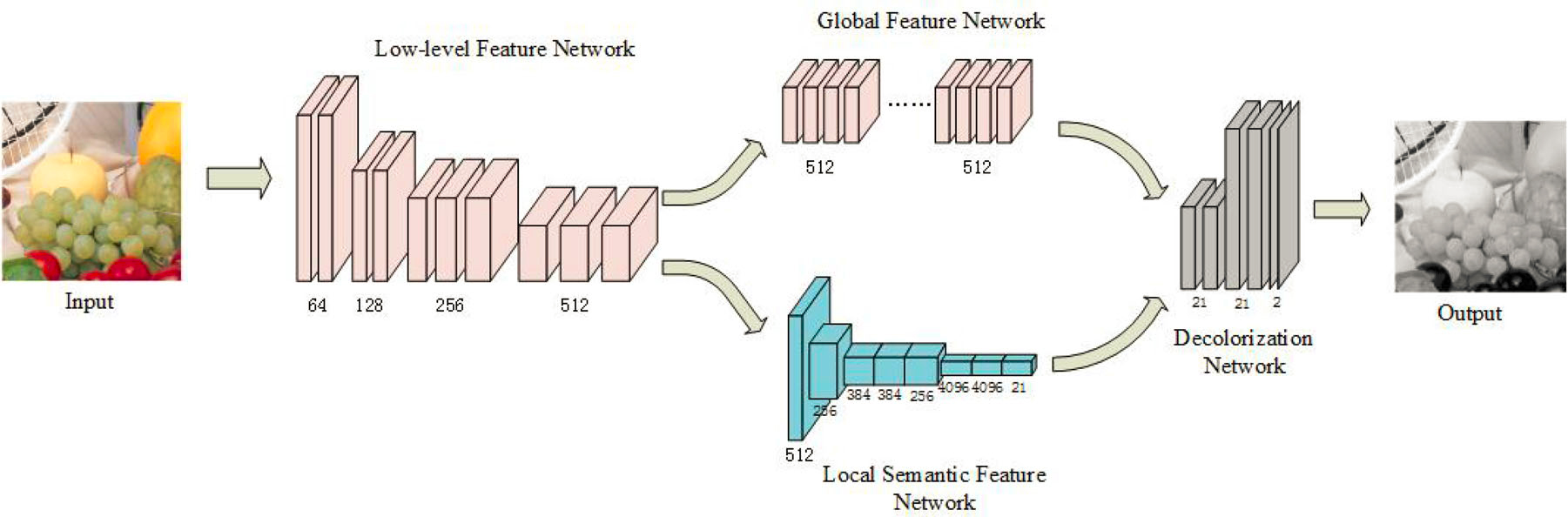}
\caption{Overview of the contrast preserving image decolorization method combining global features and local semantic features \cite{Zhang2018Decolorization}. This framework is composed of four parts: a low-level features network, a local semantic feature network, a global feature network, and a decolorization network. The four components are tightly coupled so as to learn a complex color-to-gray mapping. The low-level features network uses four groups of convolution layers to extract low-level features from the input image. With the FCN (Fully Convolutional Networks) structure, the local semantic feature network acquires instance semantic information with semantic tags of an image, such as dog and airplane. The global feature network serves to produce global image features by processing low-level features with several convolution layers. Finally, the decolorization network with the Euclidean loss outputs the resulting grayscale image.
}
\label{ch05.fig7} 
\end{figure}

Considering that the local decorization methods are less accurate enough to process local pixel leading to local artifacts, while the global methods may fail to treat local color blocks, Zhang and Liu \cite{Zhang2018Decolorization} proposed a novel image color to gray conversion method by combining local semantic features and global features. In order to preserve color contrast between adjacent pixels, a global feature network is developed to learn the global features and spatial correlation of an image. On the other hand, in order to preserve the contrast between different object blocks, they take care of local semantic features of images and fine classification of pixels during learning deep image features. Finally, with fusion of both the local semantic features and global features, this method performs better in terms of contrast preservation than the state-of-the-art decolorization approaches. Figure \ref{ch05.fig7} gives a flow chart of this method.

\begin{figure}[hbt] 
\centering
  \includegraphics[width=3.5in]{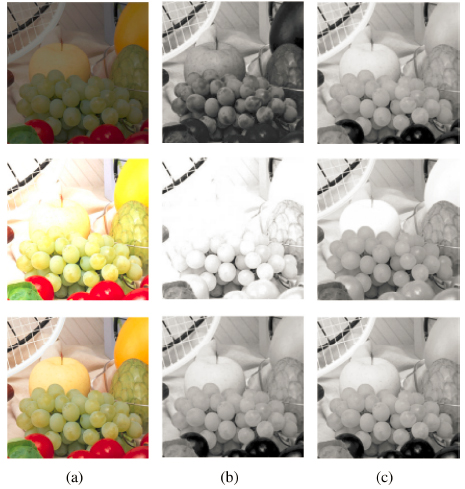}
\caption{A comparison of the results with and without the exposure feature network \cite{Liu2019Decolorization}. From left to right are input images (a), results without (b) and with (c) the exposure feature network. From top to bottom row represent low-exposure, over-exposure, and normal-exposure.}
\label{ch05.fig8} 
\end{figure}

According to the human visual mechanism, exposure plays a critical role in human visual perception, e.g., low-exposure and overexposure areas usually easily catch the attention of an observer. However, exposure is missed in existing decolorization methods. To this end, Liu and Zhang \cite{Liu2019Decolorization} proposed an image decolorization approach by fusion of local features and exposure features with a CNN framework. This framework consists of a local feature network and a rough classifier. The local feature network aims to learn the local semantic features of the color so as to maintain the contrast among different color blocks, while the rough classifier classifies three types of exposure states: low-exposure, normal-exposure, and over-exposure features of an image. Figure \ref{ch05.fig8} shows the ability of this method to treat images with different exposures.

\section{Video Colorization} 

People are willing to watch a colorful film instead of a grayscale one. \textit{Gone with the Wind} in
1939 is one of the first colorized films \cite{Gone} which is popular with the audience. However, it is challenging to obtain a convincing video colorization because of its multimodality in the solution space and the requirement of global spatiotemporal consistency \cite{Lei2019} is also inherently more challenging than Unlike single image colorization, video colorization should also satisfy temporal coherence. In view of this point, the above single image colorization cannot be used for video colorization. Currently, researches \cite{Jampani2017, Vondrick2018, Liu2018Video, Meyer2018, Lei2019}  realized video colorization by propagating the color information either from a color reference frame or sparse user scribbles to the whole target grayscale video.

Vondrick et al. \cite{Vondrick2018} regard video colorization as a self-supervised learning problem for visual tracking. To this end, they learn to colorize gray-scale videos by copying colors from a reference frame by exploiting the temporal coherency of color, rather than predicting the color directly from the gray-scale frame. Jampani et al. \cite{Jampani2017} proposed Video Propagation Network (VPN), processes video frames in an adaptive manner. The VPN consists of a temporal bilateral network (TBN) and a spatial network (SN). The TBN aims for dense and video adaptive filtering, while the SN is used for refining features and increasing flexibility. This method propagates information forward without accessing future frames. Experiments showed that, given the source color image for the first video frame, this method can propagate the color to the whole target grayscale video. Given the color image for the first video frame, the task of this method is to propagate the color to the entire video. This method can also be used for video processing tasks requiring the propagation of structured information (e.g., video object segmentation and semantic video segmentation).

\begin{figure}[t] 
\centering
  \includegraphics[width=3.5in]{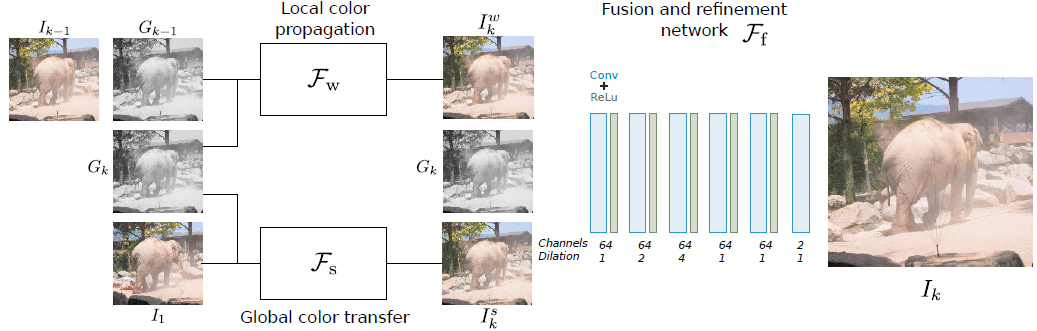}
\caption{An overview of the deep learning framework for video color propagation \cite{Meyer2018}. Both a short range network and a long range color propagation network are used to propagate colors in a video. The results of these two networks and the target grayscale image together constitute the input to the fusion and refinement network to output the final color frame.
}
\label{ch04.fig11} 
\end{figure}

Meyer et al. \cite{Meyer2018} proposed a deep learning framework for video color propagation. This method consists of a short range propagation network (SRPN), a longer range propagation network (LRPN), and a fusion and refinement network (FRN). The SRPN aims to propagate colors frame-by-frame ensuring temporal stability. The input to SRPN are two consecutive gray scale frames and it outputs an estimated warping function that is used to transfer the colors of the previous frame to the next one. 
The LRPN introduces semantical information by matching deep features extracted from the frames, which are then used to sample colors from the first frame. Except long range color propagation, this strategy also contributes to restore missing colors because of occlusion. With a CNN, the SRPN is used to combine the above two stages for fusion and refinement. Figure \ref{ch04.fig11} gives an overview of the framework of this method.

\begin{figure}[t] 
\centering
  \includegraphics[width=3.5in]{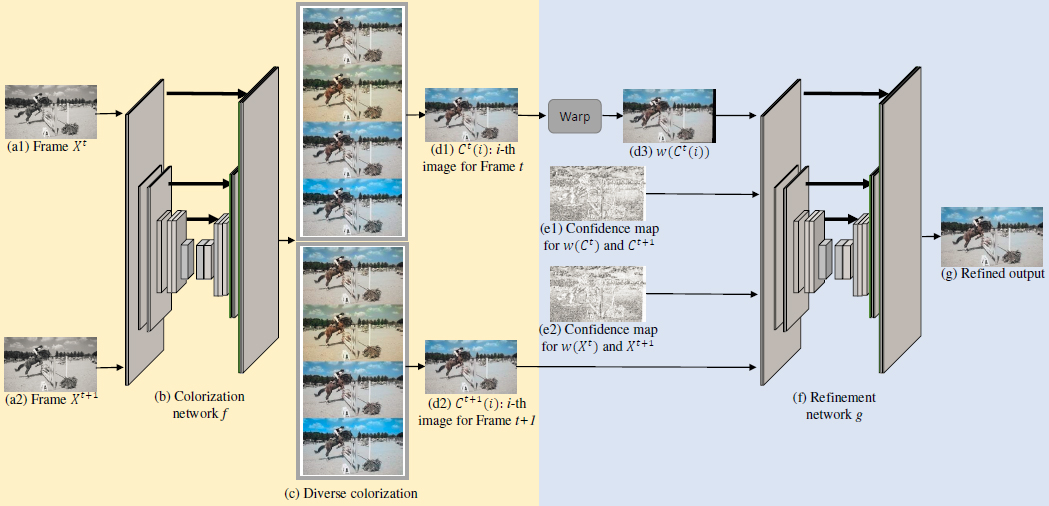}
\caption{The framework of the automatic video colorization method with self-regularization and diversity \cite{Lei2019}. This model consists of a colorization network $f$ and a refinement network $g$. $f$ is used to colorize each grayscale video frame and outputs candidate colorization images. By inputting the $i$-th colorized candidate images and two confidence maps, $g$ produces a refined video frame.}
\label{ch04.fig12} 
\end{figure}

Lei and Chen \cite{Lei2019} proposed a fully automatic, self-regularized approach to video colorization with diversity. As shown in Fig. \ref{ch04.fig12}, this method is comprised of a colorization network $f$ for video frame colorization and a refinement network $g$ for spatiotemporal color refinement. A diversity loss is designed to allow the network to generate colorful videos with diversity. Moreover, the diversity loss can also make the training and process more stable. 

\section{Video Decolorization}

As for video decolorization, people mostly extend image decolorization methods to process video frames, which would easily lead to the flicker phenomenon due to the spatiotemporal inconsistency. Video decolorization should take into account both the contrast preservation of each video frame and the temporal consistency between video frames.

Since the method of Smith et al. \cite{Smith2008} can preserve consistency avoiding changes in color ordering, they extended their two-step image grayscale transformation method to treat video decolorization.  Owing to the ability to  maintain consistency over varying palettes, Ancuti et al. \cite{Ancuti2010} applied their fusion-based decolorization technique for video cases. Given a video, Ancuti et al. \cite{Ancuti2011} searched in the entire sequence for the color palette that appears in each image (mostly identified with the static background). In this way, they extend their saliency-guided decolorization approach to video decolorization. For a video with relatively constant color palette, they computed a single offset angle value for the middle frame in a video.

Song et al. \cite{Song2014} proposed a real-time video decolorization method using bilateral filtering. Considering that human visual system is more sensitive to luminance than the chromaticity values, they recover the color contrast/detail loss in the luminance. They represent the loss as residual image by the bilateral filter. The resulting grayscale image is a sum of the residual image and the luminance of the original color image. Since the residual image is robust to temporal variations, this method can preserve the temporal coherence between video frames. Moreover, as the kernel of the bilateral filter can be set as large as the input image, this method is efficient and can run in real time on a 3.4 GHz i7 CPU.

\begin{figure}[t] 
\centering
  \includegraphics[width=3.5in]{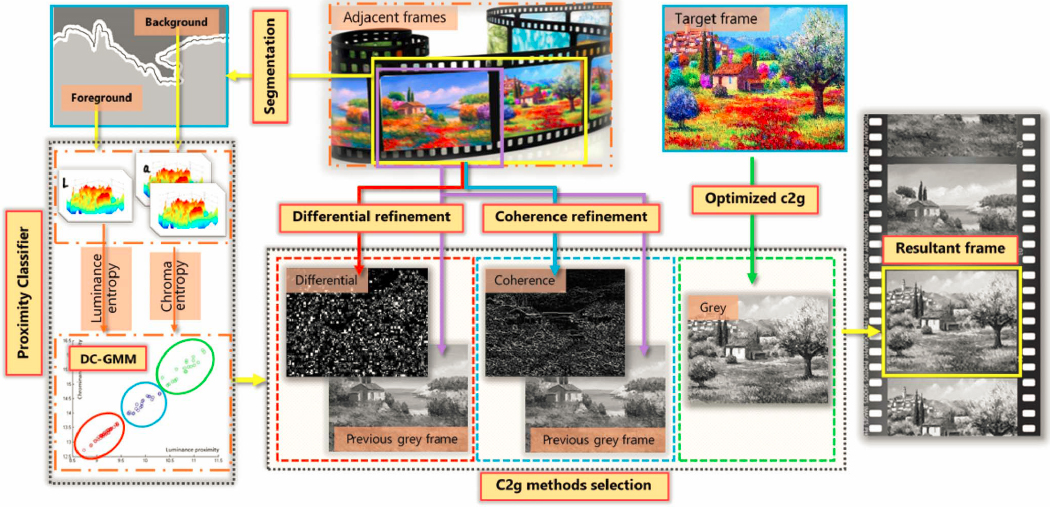}
\caption{An overview of the video decolorization using visual proximity coherence optimization \cite{Tao2017}.
Firstly, the decolorization proximity for each frame is estimated. The DC-GMM classifier is then used to select a specific decolorization strategy, and finally decolorize the frame into grayscale using the selected strategy. Secondly, with DC-GMM, video frames are classified into three categories, i.e., high-proximity, median-proximity, and low-proximity. Finally, a salience C2G method is employed to maintain temporal coherence and alleviate flickering between frames.}
\label{ch05.fig9} 
\end{figure}

Tao et al. \cite{Tao2016, Tao2017} defined decolorization proximity to measure the similarity of adjacent frames and presented a temporal-coherent video decolorization method using proximity optimization. They then respectively treat frames with low, medium, and high proximities in order to better preserve the quality of these three types of frames. Finally, with a decolorization Gaussian mixture model (DC-GMM), they classify the frames and assign appropriate decolorization strategies to them via their corresponding decolorization proximity. Figure \ref{ch05.fig9} shows an overview of this method.

\begin{figure}[t] 
\centering
  \includegraphics[width=3.5in]{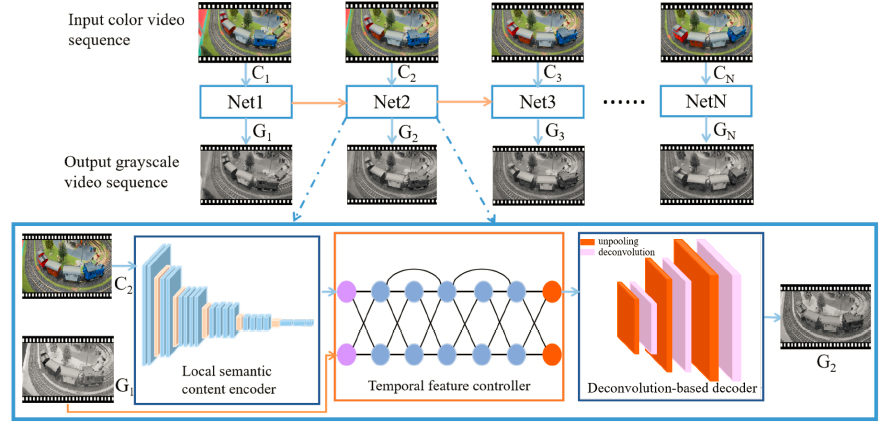}
\caption{The framework of the video decolorization method based on the CNN and LSTM neural network \cite{Liu2021Decolorization}. Given a video sequence ${C_t |t = 1, 2, 3, ..., N}$, it is processed into sequence images. Then the local semantic content encoder extracts deep features of these sequence images, adjusts the scale of the feature maps, and inputs them to the temporal features controller. After the output feature maps are fed into the deconvolution-based decoder, the resulting grayscale video sequence ${G_t |t = 1, 2, 3, ..., N}$ is produced.}
\label{ch05.fig10} 
\end{figure}

Most of the existing video decolorization methods directly apply image decolorization algorithms to treat video frames, which would easily causes temporal inconsistency and flicker phenomenon. Moreover, there may be similar local content features between video frames, which can be used to avoid redundant information. To this end, Liu and Zhang \cite{Liu2021Decolorization} introduced deep learning into the field of video decolorization by using CNN and a long short-term memory neural network. To the best of our knowledge, this is among the first attempts to perform video decolorization using deep learning techniques. A local semantic content encoder was designed to learn the same local content of a video. Here, the local semantic features were further refined by a temporal feature controller via a bi-directional recurrent neural network with long short-term memory units. Figure \ref{ch05.fig10} shows an overview of this method.

\section{Conclusion and Future Work}

This paper summarized the progress of colorization and decolorization methods for image and videos in the last two decades. According to that if user interaction is involved, we classified the image coloriztion methods into two categories, semi-automatic colorization methods and automatic colorization methods. As for image decolorization methods, we first discussed the early image decolorization methods, including the component method, the maximum method, the average method, and the weighted average method. We then summarized the existing image decolorization methods from the perspective of global decolorization and local decolorization. Finally, we also introduced the latest deep learning based colorization and decolorization approaches for images and videos.

Although convincing results can be achieved by the current colorization and decolorization methods. We think some challenges still remains. For example, a user-friendly image and video colorization and decolorization system is still needed. It is necessary to further improve the computational efficiency of the colorization and decolorization methods, especially for high-definition images and videos. Moreover, more objective metrics specific to colorization and decolorization assessment are required. Finally, large-scale datasets are needed for deep learning based image colorization and decolorization techniques. In the future, we believe researchers will pay more and more attention to this field.

\ifCLASSOPTIONcaptionsoff
  \newpage
\fi

\end{document}